\pdfoutput=1

\documentclass[11pt]{article}

\usepackage{latex/acl}

\usepackage{times}
\usepackage{latexsym}
\usepackage{amssymb}
\usepackage{pifont}
\usepackage{xcolor}
\usepackage{mdframed}

\usepackage[T1]{fontenc}

\usepackage[utf8]{inputenc}

\usepackage{microtype}

\usepackage{inconsolata}
\usepackage{graphicx}
\usepackage{dblfloatfix}
\usepackage{siunitx} 
\usepackage{booktabs} 
\usepackage{tabularx} 
\usepackage{booktabs}  
\usepackage{multirow} 
\usepackage{siunitx} 

\title{OSCaR: Object State Captioning and State Change Representation}

\author{Nguyen Nguyen\textsuperscript{1}, Jing Bi\textsuperscript{1}, Ali Vosoughi\textsuperscript{1}, Yapeng Tian\textsuperscript{2}, Pooyan Fazli\textsuperscript{3}, Chenliang Xu\textsuperscript{1} \\
  \textsuperscript{1}University of Rochester, 
  \textsuperscript{2}University of Texas at Dallas, 
  \textsuperscript{3}Arizona State University \\
  \texttt{\{nguyen.nguyen, jing.bi, ali.vosoughi, chenliang.xu\}@rochester.edu}, \\
  \texttt{yapeng.tian@utdallas.edu}, \texttt{pooyan@asu.edu}}

\begin{document}

\maketitle

\begin{abstract}
The capability of intelligent models to extrapolate and comprehend changes in object states is a crucial yet demanding aspect of AI research, particularly through the lens of human interaction in real-world settings.
This task involves describing complex visual environments, identifying active objects, and interpreting their changes as conveyed through language. 
Traditional methods, which isolate object captioning and state change detection, offer a limited view of dynamic environments. 
Moreover, relying on a small set of symbolic words to represent changes has restricted the expressiveness of language.
To address these challenges, in this paper, we introduce the Object State Captioning and State Change Representation (OSCaR) dataset and benchmark. 
OSCaR consists of 14,084 annotated video segments with nearly 1,000 unique objects from various egocentric video collections. 
It sets a new testbed for evaluating Multimodal Large Language Models (MLLMs). 
Our experiments demonstrate that while MLLMs show some skill, they lack a full understanding of object state changes. 
The benchmark includes a fine-tuned model that, despite initial capabilities, requires significant improvements in accuracy and generalization ability for effective understanding of these changes. Our code and dataset are available at \url{https://github.com/nguyennm1024/OSCaR}.

\end{abstract}

\section{Introduction}


The field of Natural Language Processing (NLP) has evolved beyond mere text interpretation and generation, advancing into realms where understanding and interacting with the physical world becomes imperative. 
\begin{figure}[htbp]
\centering
\includegraphics[width=0.8\columnwidth]{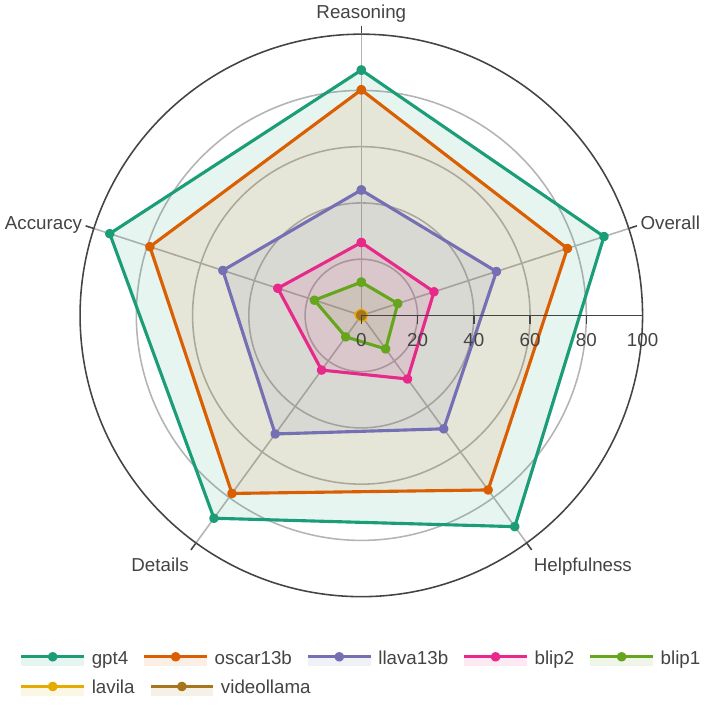}
\caption{\textbf{Surpassing prior models in aligning with human judgements.} Our method achieves near parity with GPT-4V ratings across helpfulness, accuracy, reasoning, and other key metrics. }
\label{fig:compare}
\end{figure}
From studying causal reasoning~\cite{gao2018action} to building a world model for cause-effect prediction~\cite{gao2016physical,alayrac2017joint}, researchers have been working on the problem of causation in the physical world.

\begin{figure*}[t!] 
    \centering
    \captionsetup{type=figure}
    \includegraphics[width=0.9\textwidth, height=0.35\textwidth]{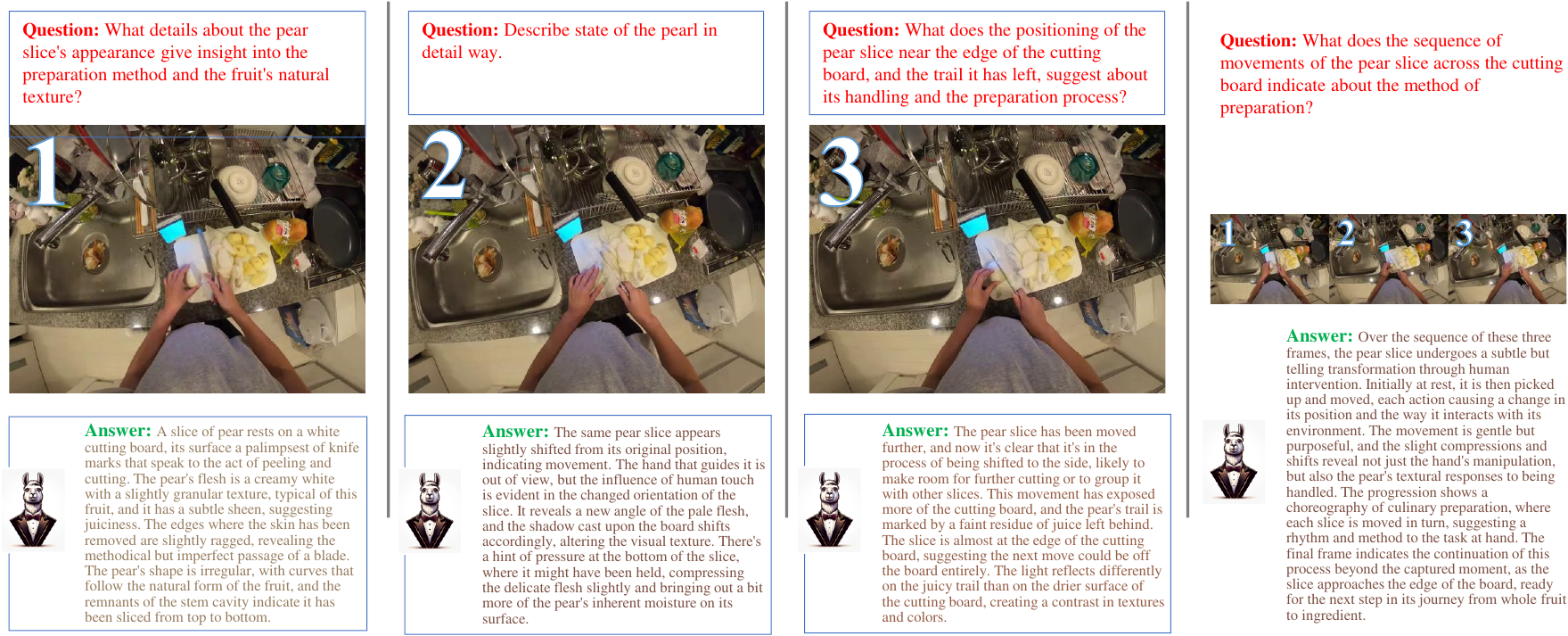}
    \caption{\textbf{OSCaR's description of state, state change, and illustration of reasoning.} State description involves the characterization of a specific region of interest within the video and the associated activity. State change entails the description of the evolution of a system over a defined temporal sequence. Furthermore, the analysis of the state of an object is centered on comprehending and elucidating the mechanisms underlying the object's evolution.}
    \label{fig:oscar-example}
\end{figure*}

In this paper, we investigate the very basic causal relations between a concrete action and the change of the object state caused by this action. 
For example, given an image as shown in Figure~\ref{fig:oscar-example}, we, as humans, would have no problem understanding which object is being actively interacted with.
Furthermore, given the statement "cutting the bread", we would naturally imagine what state change may happen.
However, Despite tremendous progress in knowledge representation, automated reasoning, and machine learning, artificial agents still lack the understanding of naive causal relations regarding the physical world \cite{gao2018action}.

Imagining a scenario where artificial agents collaborate with humans in the physical world, they will need to understand the physical action effect to reason, learn, and assist humans~\cite{Bi2023MISARAM}.
To empower machines with such capabilities, this paper introduces a novel benchmark focusing on understanding object state changes from egocentric visual inputs, which has the advantage of the lens of human eyes.

Understanding object state change is not only a complex task but also practical and foundational for many other tasks, such as helping intelligent agents to understand the environment dynamics and complete task~\cite{padmakumar2023multimodal,sarch2023open,merullo2022pretraining}, tracking the state of dialog\cite{le2022multimodal}, creating causal graphs for knowledge representation for complex question and answering~\cite{ates2020craft}.

Modeling object state change requires two abilities: 1) scene understanding, which involves parsing the world through an object-centric lens, and 2) causal-effect understanding, which entails identifying likely actions and their effects by observing images before, during, and after an action.

Previous research efforts have concentrated on building symbolic representations to ground changes and states~\cite{wu2023localizing,zellers2021piglet,nagarajan2018attributes}. However, given the diversity and complexity of objects and their states, influenced by contextual and temporal factors, symbolic representation alone falls short.
This paper proposes the use of natural language as a more expressive and intuitive medium for this task. This approach not only aligns the understanding of visual content between humans and AI systems but also enhances communication between them, providing a richer context than unimodal models.

Essentially, we form the scene understanding as an object-centric visual captioning problem. 
We can utilize natural language to describe the objects and any changes that may occur. 
On the other hand, the ability to understand the causal effect is formed as a visual question-answering problem based on 3 images: before, during, and after the action. Our dataset and experiments exhibit considerable potential for scalable application across various domains in future research. While conducting this study, another research was also conducted to understand object state change with a different approach~\cite{xue2024learning}. That shows the importance and significant potential of this research direction.

In summary, our contributions are threefold:
\begin{itemize}
    \item We introduce a new problem to understand states and state changes of object through natural language.     \item We present a method to generate good-quality visual instructions guided by simple annotations, applicable to both images and videos, advancing future research in visual instruction tuning. Our pipeline provides a good starting point for the data collection process.
    \item Our paper introduces OSCaR, a novel dataset and a benchmark leveraged by the power of GPT-4V that contains different tasks for object state understanding, including visual captioning, visual question answering visual dialog, and reasoning.
\end{itemize}

\section{Related Works}
\noindent \textbf{Object state change:}
Localizing and recognizing changes of object states, play a key role in applications such as procedural planning~\cite{Bi_2021}, robotics, and video action understanding~\cite{du2023learning,zhong2023learning,tang2023llmva,wang2023manipulate,songaudio}.
Recognizing object state changes necessitates the joint discovery of states and actions through an understanding of their causal relationship, as discussed in prior works ~\cite{alayrac2017joint,liu2017jointly,souvcek2022look,naeem2021learning}. 
Recently, a self-supervised method has been proposed to jointly localize action and state changes temporally from noisy untrimmed long videos~\cite{souvcek2022look}.
Moreover, ~\cite{saini2023chop} introduces a novel benchmark for the generation of object states, yet their focus is very limited to only the cutting action and a small dataset.
However, previous studies often separate scene understanding from object state change recognition and tend to operate under a closed-world assumption, which limits their applicability in real-world scenarios.
Our research aims to bridge the gap between human and machine perception by integrating egocentric views and language. 

\noindent \textbf{Multimodal Large Language Models:} Recent advancements in  Large Language Models (LLMs)~\cite{ouyang2022training_instructGPT, touvron2023llama, chiang2023vicuna, chung2022scaling_flant5} have led to significant achievements in language understanding and generation.
This progress has sparked an interest in the creation of MLLMs that blend the advanced linguistic processing of LLMs with capabilities for multi-modal perception~\cite{2023vpgtrans, ye2023mplugowl, li2023otter, gao2023llamaadapterv2, peng2023kosmos,tang2023video}. 
The core of this research is the fusion of pre-training visual encoder representations with the input embedding space of LLMs, achieved by pretraining with datasets that interleave images and text. \cite{Li2023BLIP2BL, zhu2023minigpt, liu2023visual_llava}. 
In this paper, we aim to provide a comprehensive evaluation of these models, particularly focusing on their performance in object state change recognition.

\section{The OSCaR Dataset}
This section outlines our pipeline for creating visual instructions on object states. We begin with the process of collecting diverse visual data from public sources, detailed in section~\ref{sec:video-collection}. Following this, section~\ref{sec:gpt-assist} describes our approach to enhancing data quality using simple human annotations across various tasks, facilitating a deeper understanding of object states. Our method enables the generation of detailed captions, visual question answering, and visual dialogue.

\subsection{Video Collections}
\label{sec:video-collection}

OSCaR is a curated compilation of videos sourced from two distinct datasets: EPIC-KITCHENS~\cite{Damen2018ScalingEV} and Ego4D~\cite{Grauman2021Ego4DAT}. Acknowledging that changes in object states occur progressively over time rather than abruptly within a single frame, we have selectively included video clips that effectively illustrate these state transitions. Our selection process ensures that these videos depict the dynamic changes in object states and capture moments where the objects remain stationary for short enough durations. This approach enabled us to compile a comprehensive visual dataset encompassing the object's static and transitional states.

We initially analyzed the verbs from the original videos of the EPIC-KITCHENS dataset to ensure that the videos highlighted objects undergoing state changes. We categorized these verbs into three groups: \textit{change}, \textit{not sure}, and \textit{not change}. The \textit{change} group consists of verbs likely to alter the state of objects, including actions like Open, Close, Wash, Cut, and Mix. Conversely, the \textit{not change} group encompasses verbs with a minimal likelihood of inducing state changes, such as Take, Put, Move, Check, etc. Lastly, the \textit{not sure} group includes verbs with ambiguous potential for state change, covering actions like Shake, Flip, Use, Pull, and others. After filtering the EPIC-KITCHENS dataset, we were able to identify 69 verb classes that consisted of a total of 650 verbs. Using this verb list, we retrieved all video segments containing those actions.

Upon analyzing the videos, we discovered that some objects only appeared in a few times. As a result, we split the videos into two groups. The first group comprises videos that focus on objects that occurred more than ten times, and it will be used to construct our training and testing set. The second group includes videos with objects that occurred less than ten times. These objects are rare in EPIC-KITCHENS and can be used for open-world evaluation, which will be discussed in section~\ref{sec:open-world}. In the first group, we randomly selected 10 to 50 video segments per object, resulting in 7442 with 306 different objects from EPIC-KITCHENS.

We leveraged Ego4D, the largest egocentric video dataset, selecting video segments tagged with "$object\_of\_change$" to enhance our data's diversity. This tag highlighted videos showcasing object state changes. By gathering these specific videos, along with details of the objects and their narrations, we informed our data generation and compiled relevant statistics. From this dataset, we extracted 5942 segments featuring 296 unique objects for our OSCaR project.

\subsection{GPT-assisted Data Generation}\label{sec:gpt-assist}
\textbf{Caption Generation:} Captioning plays an important role in visual understanding. Understanding object states requires detailed and informative captions to capture the exact state of objects. To achieve this goal, we generated captions for all collected videos by leveraging GPT-4V and human's weak annotations. This problem requires two types of annotations, including 1) Start and end frame ID in videos during the event to make state changes and 2) A short description of what happens in the video. The short description can be a verb representing the action and a noun representing the object humans interact with (e.g., washing tray). We designed adaptive prompts to inject this annotation as context to guide GPT-4V to generate high-quality captions. We found that GPT-4V often suffers from ambiguity without this guidance, and the quality of generated captions is degraded. With simple human guidance, GPT-4V can reduce ambiguity and produce better-quality captions.\newline
\noindent \textbf{Multiple-choice QA Generation:} The multiple-choice question is a method of presenting a set of answers, including incorrect options, to teach machine learning models how to distinguish between correct and incorrect answers. This type of question can also be used as a form of instruction, where the question serves as the prompt, and the answer serves as the response for the models. We created multiple-choice question and answer sets based on generated captions.\newline
\noindent \textbf{Conversation Generation:} Visual dialog is a complex task requiring understanding of visual content and conversation context, and faces challenges in data collection due to its need for natural dialogues between two people viewing the same content. This process is time-consuming and resource-intensive, especially when involving reasoning and explanations. With the growth of machine learning models, generating visual dialog data is increasingly vital. We've developed a method that uses captions to create visual conversation data, enhanced by GPT-4V's ability to provide explanations, offering flexible and diverse data. This approach, labeling input data for images and videos, is cost-effective and faster than manual methods, generating vast amounts of training data for future models.

\section{OSCaR Benchmarks}
\subsection{Evaluation with Text Generation Metrics}
The dataset we are providing consists of 500 videos from the Ego4D and EPIC-KITCHENS datasets, which are specifically designed for benchmarking purposes. Each video is annotated by four detailed captions, all of which have undergone rigorous human verification to ensure the quality and reliability of this evaluation set. To ensure a comprehensive and accurate assessment of performance, text generation metrics such as BLEU, Rouge, LSA, among others, can be used for evaluation purposes.

\subsection{Open-world Object State Understanding}
\label{sec:open-world}
Collecting data for all objects worldwide and then training models is not feasible. However, humans can describe new or unfamiliar objects, which can be challenging for AI, especially when they are in a new domain or serve a different purpose. Fortunately, recent achievements in MLLMs have opened up the potential for AI to have this ability. During pre-training with large amounts of data, MLLMs can learn general knowledge about the world. Besides, models will learn how to perform tasks during the visual instruction tuning process. In both processes, the models may or may not have been exposed to objects not in the object state understanding training set. The question is whether models can generalize to objects of this type. To answer this question, we provide two evaluation sets to test the generalizability of the models.

\noindent \textbf{Cooking domain objects have not occurred in the training set for object state understanding:} For this evaluation, we want to investigate the model's ability to understand objects that have not appeared in the training set in a similar scenario with the training domain. We provided a set of 2,485 videos with 1,024 objects that have not occurred in the object state training set. This testing set will evaluate how in-domain knowledge can help models understand object states and state changes. We used GPT-4V to annotate 344 videos for evaluation purposes.

\noindent \textbf{Out-of-domain objects state understanding:} This evaluation focuses on judging the ability of models to understand objects beyond the training domains. Our training set contains only the cooking domain data, while this testing set has diverse domains, such as baker, household management, cleaning/laundry, bike mechanic, etc. This set was extracted from the Ego4D dataset and contains 43,367 videos with more than 500 objects. This testing not only can be used for evaluation but also has the potential to scale up using our pipeline for object state understanding in other specific domains. For this evaluation set, we selected 10 videos from each of the 51 different domains, totaling 356 videos. Domains with fewer than 10 videos have all their videos included. This set is also annotated by GPT-4V.

\subsection{Data Quality Verification}
\label{sec:quality-verification}

We evaluated the quality of descriptions for object states and activities across video frames using Amazon MTurk for human feedback. Our assessment framework included five guidelines for spotting inaccuracies, focusing on frame-specific description accuracy, two for assessing state change accuracy, two for identifying hallucinations, and three for recognizing incomplete descriptions. Annotators were asked to categorize each description under one of four labels: 1) Fully Detailed and Comprehensive, 2) Generally Complete with Minor Omissions, 3) Lacks Important Details or Contains Errors, or 4) Incomplete, Misleading, or Hallucinating, and provide reasoning to discourage random responses. This study utilized 500 samples from the EPIC-KITCHENS and Ego4D datasets, leading to the validation of 2000 natural language descriptions.

\begin{figure}[!t]
\centering
\includegraphics[width=\columnwidth]{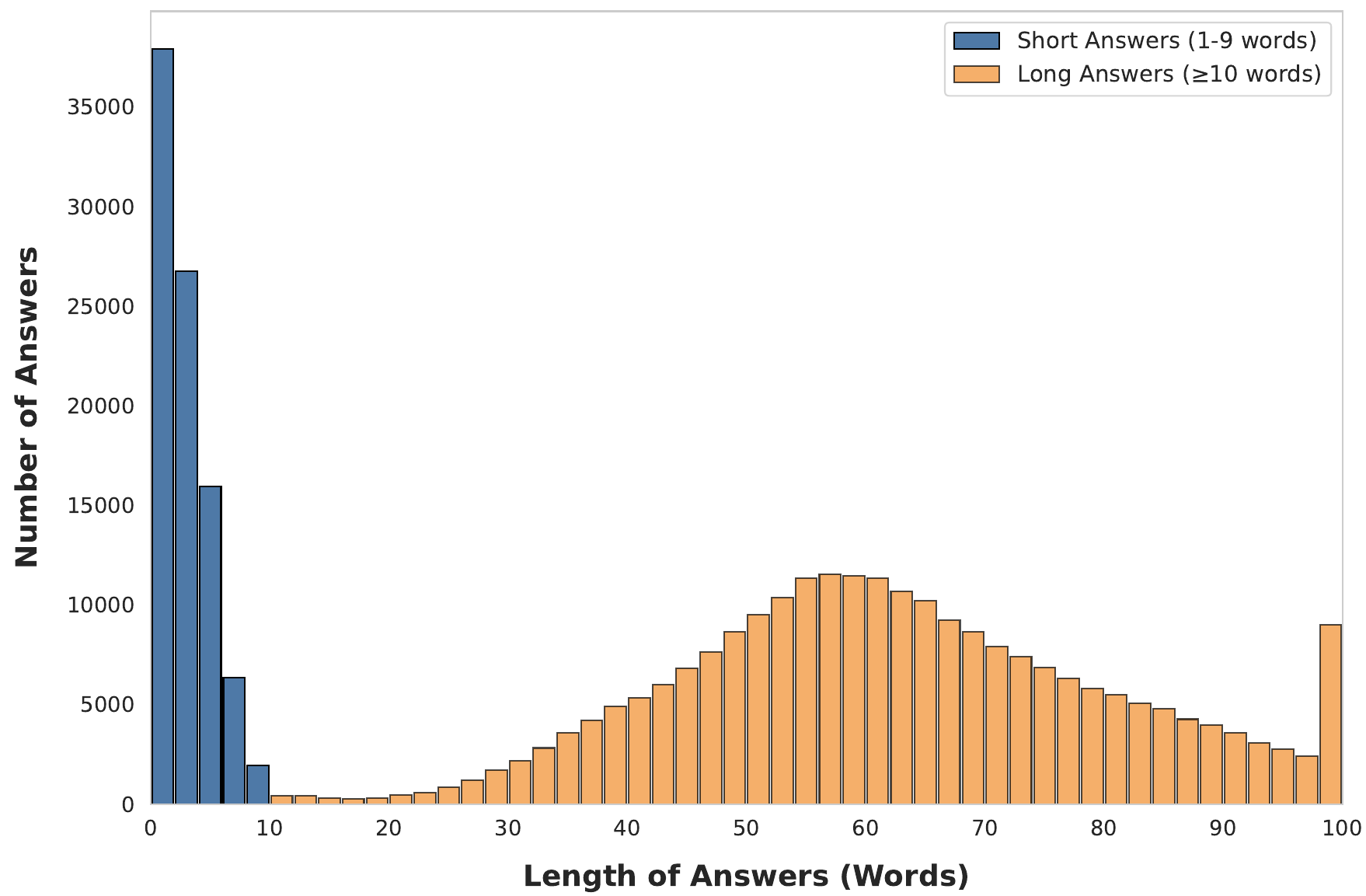}
\caption{\textbf{Distribution of answer lengths.} The figure shows how answers are distributed by length in the dataset. It separates short answers (1-9 words) from long answers ($\ge 10$ words). The histogram displays the number of answers on the y-axis based on increasing answer lengths on the x-axis. There is a category at 100 words for answers with lengths greater than or equal to 100 words. This breakdown emphasizes the balance between brief, direct answers and more detailed, explanatory responses.}

\label{fig:ans-len}
\end{figure}
\section{Data Statistics}

\begin{figure}[!t]
\centering
\includegraphics[width=\columnwidth]{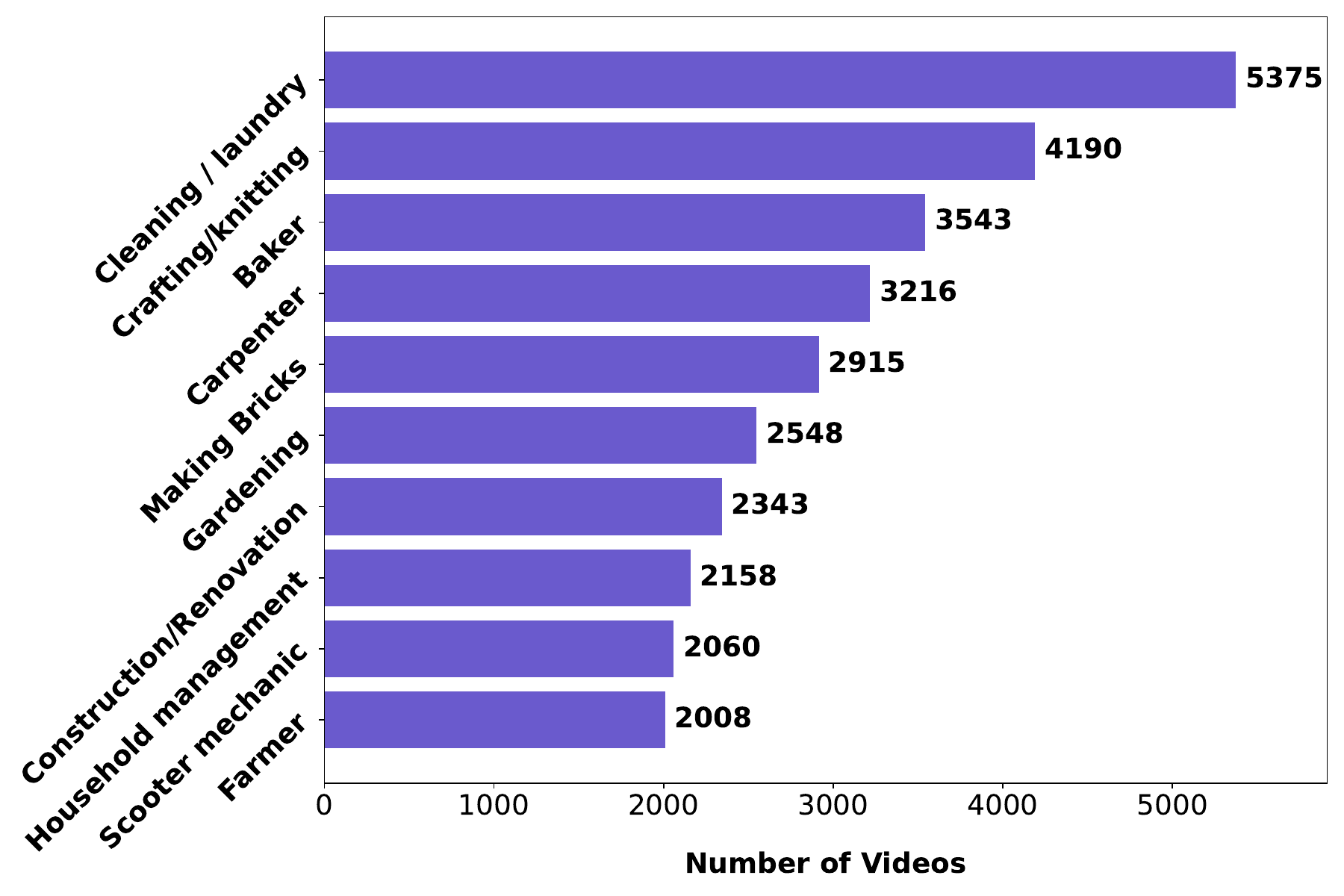}
\caption{\textbf{Top 10 open-world domains (excluding cooking)}. The figure shows non-cooking domains present in the open-world test set used to assess model generalization. By evaluating performance on household and occupational activities unseen during training, we benchmark the trained models' capacity to understand new objects and actions beyond cooking tasks. }
\label{fig:ow-domain-dist}
\end{figure}

In order to help models generate concise and informative answers, we have defined short answers as those with less than ten words and long answers as those with more than ten words. Short answers provide brevity, while long answers offer detailed and informative information. The distribution of these two types of answers can be seen in Figure~\ref{fig:ans-len}. The average answer length in the dataset is 47.06 words. Long answers make up about 75\% of the data, with an average length of 63 words, while short answers account for about 25\% of the data, with an average length of 3.32 words. By splitting the data accordingly, future models can provide short, direct, and informative answers with explanations. To showcase the uniqueness of our OSCaR dataset, we have presented a comparison between OSCaR and other related datasets in Table~\ref{tab:datasets}. The OSCaR dataset comprises a vast number of instructions, along with images and videos. Additionally, it also provides data for object state captioning and object state change captioning.

In section~\ref{sec:open-world}, we discussed two types of open-world datasets for object state understanding: in-domain cooking and open domains. Although we trained on videos with object state changes, in open-world evaluation, we tested the models on both types of videos, with and without object state changes, to ensure their generalizability. The in-domain evaluation set consists of 2,485 videos with 1,024 novel objects extracted from EPIC-KITCHENS. 

We have extracted an open-domain evaluation set from the Ego4D dataset. The top 10 most frequent domains in the open-world testing set are shown in Figure~\ref{fig:ow-domain-dist}. This evaluation set from 51 different domains, contains annotations for domain, action, object name, and action narrations extracted from annotations of Ego4D. The set includes 43,367 open-world videos for which we know their domains and 56,231 videos of unknown domains, but we still have information about their object names and action narrations. Thus, this set can be utilized not only for open-world evaluation but also for the advancement of general domain object-state understanding in the future when applying our method to generate labels. This set of data hasn't been annotated, but the data we extracted from Ego4D are ready to use our pipeline to scale up the data generation.

\newcommand{\boldcheckmark}{\scalebox{1.2}{$\checkmark$}}

\begin{table*}[ht]
\centering
\caption{Comparison of OSCaR dataset versus other related datasets. OSC and OSCC represented for Object State Captioning and Object State Change Captioning, respectively.}
\label{tab:datasets}
\scalebox{0.75}{
\setlength{\tabcolsep}{15pt} 
\begin{tabular}{@{}lccccc@{}} 
\toprule
\textbf{Dataset} & \textbf{Video} & \textbf{\#Clip} & \textbf{\#Instruction} & \textbf{OSC} & \textbf{OSCC} \\
\midrule
MiniGPT-4 \cite{Zhu2023MiniGPT4EV} & {\color{gray}\ding{55}} & {\color{gray}\ding{55}} & 5K & {\color{gray}\ding{55}} & {\color{gray}\ding{55}} \\
Shikra-RD \cite{Chen2023ShikraUM} & {\color{gray}\ding{55}} & {\color{gray}\ding{55}} & 5.9K & {\color{gray}\ding{55}} & {\color{gray}\ding{55}} \\
LLaVA \cite{Liu2023VisualIT} & {\color{gray}\ding{55}} & {\color{gray}\ding{55}} & 345K & {\color{gray}\ding{55}} & {\color{gray}\ding{55}} \\
VideoChat \cite{Li2023VideoChatCV} & \boldcheckmark & 11K & 20.8K & {\color{gray}\ding{55}} & {\color{gray}\ding{55}} \\
OSCaR & \boldcheckmark & \textbf{18K} & \textbf{400K} & \boldcheckmark & \boldcheckmark \\
\bottomrule
\end{tabular}
}
\end{table*}

\section{Experiments}

In this section, we will discuss the experimental design we used and how we trained our model. Our fine-tuning process will be described in Section~\ref{sec:training}. Additionally, we included other vision language models such as BLIP~\cite{li2023blip} , BLIP2~\cite{Li2023BLIP2BL}, LaViLa~\cite{Zhao2022LearningVR}, and Video-LLaMA~\cite{Zhang2023VideoLLaMAAI} for comparison purposes. Firstly, we will evaluate our model's performance in the cooking domain in Section~\ref{sec:cooking-eval}. After that, we will also evaluate its performance in an open-world setting in Section~\ref{sec:open-world-eval}.


\subsection{Model Training}
\label{sec:training}

We conducted extensive experiments to showcase the effectiveness of our data generation pipeline in solving object-state understanding problems. A straightforward approach to solving these types of problems is using a model with a text encoder to encode prompts and a visual encoder to encode visual content. After that, both of these inputs will be used as conditions to generate text answers with a text decoder. Ideally, this text decoder will be an LLM.

We fine-tuned LLaVA, an open-source MLLM featuring capabilities like visual dialogue, question-answering~\cite{Agrawal2015VQAVQ}, and OCR~\cite{Nguyen2021DictionaryguidedST, Nguyen2024EfficientlyLL}, to achieve our goals. Notably, the generated data can enhance any future vision-language models beyond LLaVA. We experimented with LLaVA using Vicuna 7B and 13B models under two conditions: with and without its original visual instruction tuning data, referring to the former as OSCaR.

For training, we employed Lora fine-tuning with a configuration of rank 128 and alpha 256, using Vicuna 13B and 7B models alongside the OpenAI/CLIP-ViT-Large-Patch14-336 vision encoder. A projector transformed visual features into tokens. Our fine-tuning parameters included a single epoch, a learning rate of 2e-4, a batch size of 16 per device, and a maximum model length of 2048.

\subsection{Evaluating GPT-4V}
\begin{figure}[!t]
\centering
\includegraphics[width=0.9\columnwidth]{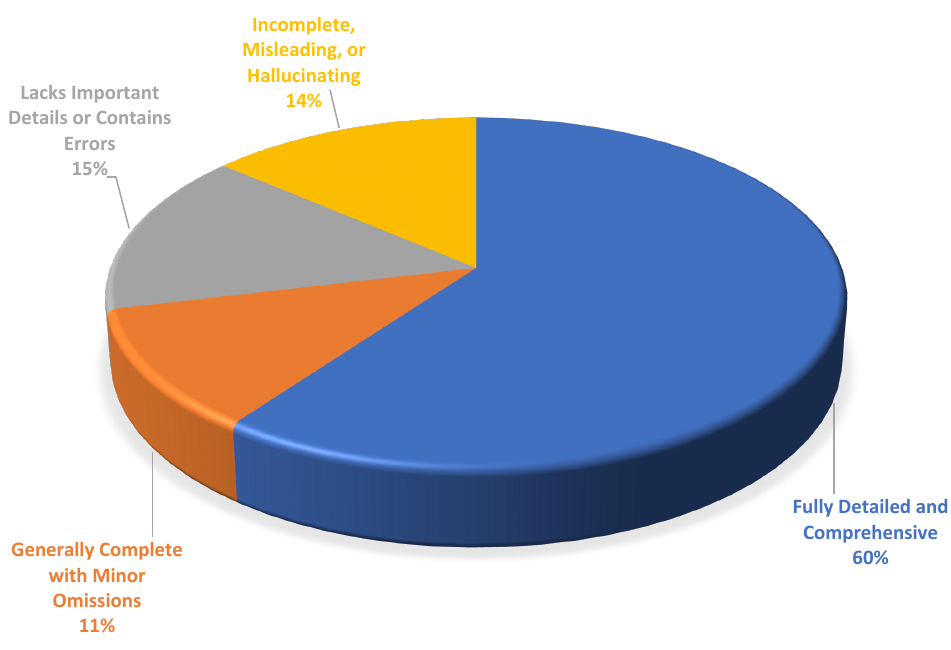}
\caption{\textbf{GPT-4V zero-shot caption quality human evaluation.} The figure shows the distribution of quality ratings assigned by human annotators evaluating frame descriptions automatically generated by the GPT-4V model under zero-shot conditions. Descriptions for 500 video frames were rated.}
\label{fig:gpt4-zeroshot}
\end{figure}

Because our pipeline uses GPT-4V as the knowledge model to annotate our data, evaluating GPT-4V's ability is crucial. Evaluating GPT-4V's performance has two purposes: 1) Understanding the performance of GPT-4V on this task and 2) Producing a clean benchmark beyond the ability of GPT-4V for future research. As discussed in section~\ref{sec:quality-verification}, we ask humans to check data quality and classify quality into four levels with text explanation. Figure~\ref{fig:gpt4-zeroshot} shows the distribution of data quality from 500 videos sampled from the dataset for benchmarking.

\begin{table*}[ht]
\centering
\caption{\textbf{Performance comparison based on BLEU and ROUGE scores.} OSCaR is LLaVA fine-tuned with OSCaR data, mixed data is a combination of LLaVA data and OSCaR data.}
\label{tab:model_performance}
\scalebox{0.65}{
\begin{tabular}{@{}lSSSS@{}} 
\toprule
\textbf{Model} & {\textbf{BLEU}} & {\textbf{ROUGE-1}} & {\textbf{ROUGE-2}} & {\textbf{ROUGE-L}} \\
\midrule
LaViLa~\cite{Zhao2022LearningVR} & 0.006 & 3.3 & 0.26 & 3.27 \\
BLIP1~\cite{Li2022BLIPBL} & 0.008 & 1.38 & 0.08 & 1.35 \\
BLIP2~\cite{Li2023BLIP2BL} & 0.1 & 11.53 & 2.12 & 10.51 \\
Video-LLaMA~\cite{Zhang2023VideoLLaMAAI} & 1.0 & 17.75 & 2.69 & 16.02 \\
LLaVA v1.5 13B~\cite{Liu2023VisualIT} & 3.72 & 27.09 & 6.59 & 24.01 \\
LLaVA v1.5 7B~\cite{Liu2023VisualIT} & 3.23 & 25.37 & 6.22 & 22.60 \\
\midrule
OSCaR 13B (OSCaR data only) (Ours) & 5.28 & 27.93 & 7.67 & 24.45 \\
OSCaR 7B (OSCaR data only) (Ours) & 5.1 & 28.27 & 7.42 & 24.77 \\
OSCaR 13B (Mixed data) (Ours) & 5.76 & 29.26 & 8.24 & 25.78 \\
OSCaR 7B (Mixed data) (Ours) & \textbf{5.79} & \textbf{29.94} & \textbf{8.34} & \textbf{26.24} \\
\bottomrule
\end{tabular}
}
\end{table*}

\subsection{Evaluation on Cooking Domain Objects}
\label{sec:cooking-eval}
\begin{table*}[ht]
\centering
\caption{\textbf{Open-world performance comparison based on BLEU and ROUGE scores.} OSCaR is LLaVA fine-tuned with OSCaR data, mixed data is a combination of LLaVA data and OSCaR data.}
\label{tab:openworld_performance}
\scalebox{0.7}{
\begin{tabularx}{\textwidth}{@{}lXcccc@{}}
\toprule
\textbf{Open World} & \textbf{Model} & \textbf{BLEU} & \textbf{ROUGE-1} & \textbf{ROUGE-2} & \textbf{ROUGE-L} \\
\midrule
\multirow{4}{*}{In Domain} 
& OSCaR 13B (OSCaR data only) & 5.86 & 28.64 & 8.43 & 24.91 \\
& OSCaR 7B (OSCaR data only) & 5.73 & 29.10 & 8.38 & 25.47 \\
& OSCaR 13B (Mixed data) & \textbf{6.19} & 29.36 & 8.74 & 25.69 \\
& OSCaR 7B (Mixed data) & 6.13 & \textbf{30.00} & \textbf{8.95} & \textbf{26.25} \\
\midrule
\multirow{4}{*}{Out of Domain} 
& OSCaR 13B (OSCaR data only) & 5.32 & 27.20 & 7.62 & 23.67 \\
& OSCaR 7B (OSCaR data only) & 5.18 & 27.07 & 7.50 & 23.65 \\
& OSCaR 13B (Mixed data) & 5.24 & 26.18 & 7.36 & 23.09 \\ 
& OSCaR 7B (Mixed data) & \textbf{5.69} & \textbf{28.99} & \textbf{8.29} & \textbf{25.38} \\
\bottomrule
\end{tabularx}
}
\end{table*}
\noindent \textbf{Text Generation Metrics Evaluation:} The table~\ref{tab:model_performance} in this document displays the results of two text generation metrics, BLEU and ROUGE. As per the table, LaViLa and BLIP1 models have scored very low, whereas BLIP2, Video-LLaMA, and LLaVA models, which are currently the most advanced models, have achieved significant improvements. Our proposal has surpassed every previous state-of-the-art model by a large margin on these metrics.\\
\noindent \textbf{GPT4 Evaluation:} The experimental results of evaluating LLaVA, OSCaR, and GPT-4V captions on five criteria using GPT-4V are shown in Table~\ref{tab:gpt4-eval}.
\begin{table}[h]
\centering
\caption{Evaluation scores using GPT-4V under different criterion are listed in the table.}
\label{tab:gpt4-eval}
\scalebox{0.7}{
\begin{tabularx}{\columnwidth}{XXXX}
\toprule
\textbf{Criteria} & LLaVA & OSCaR & GPT-4V \\
\midrule
\textbf{Accuracy} & 53.60 & 82.93 & 94.04 \\
\textbf{Helpfulness} & 51.63 & 80.78 & 92.83 \\
\textbf{Reasoning} & 53.64 & 79.20 & 87.22 \\
\textbf{Detail} & 40.56 & 87.30 & 89.14 \\
\textbf{Overall} & 51.96 & 80.92 & 90.72 \\
\bottomrule

\end{tabularx}
}
\end{table}
According to the metric used, OSCaR performs significantly better than LLaVA. Additionally, OSCaR achieved 88.19\%, 87.01\%, 90.81\%, 89.21\%, and 97.94\% in accuracy, helpfulness, detail level, reasoning, and overall, respectively, compared to GPT-4V. On average, OSCaR is \textbf{90\%} as good as GPT-4V. The visualization can be seen at Figure~\ref{fig:compare}.\\

\begin{figure}[!t]
\centering
\includegraphics[width=0.9\columnwidth]{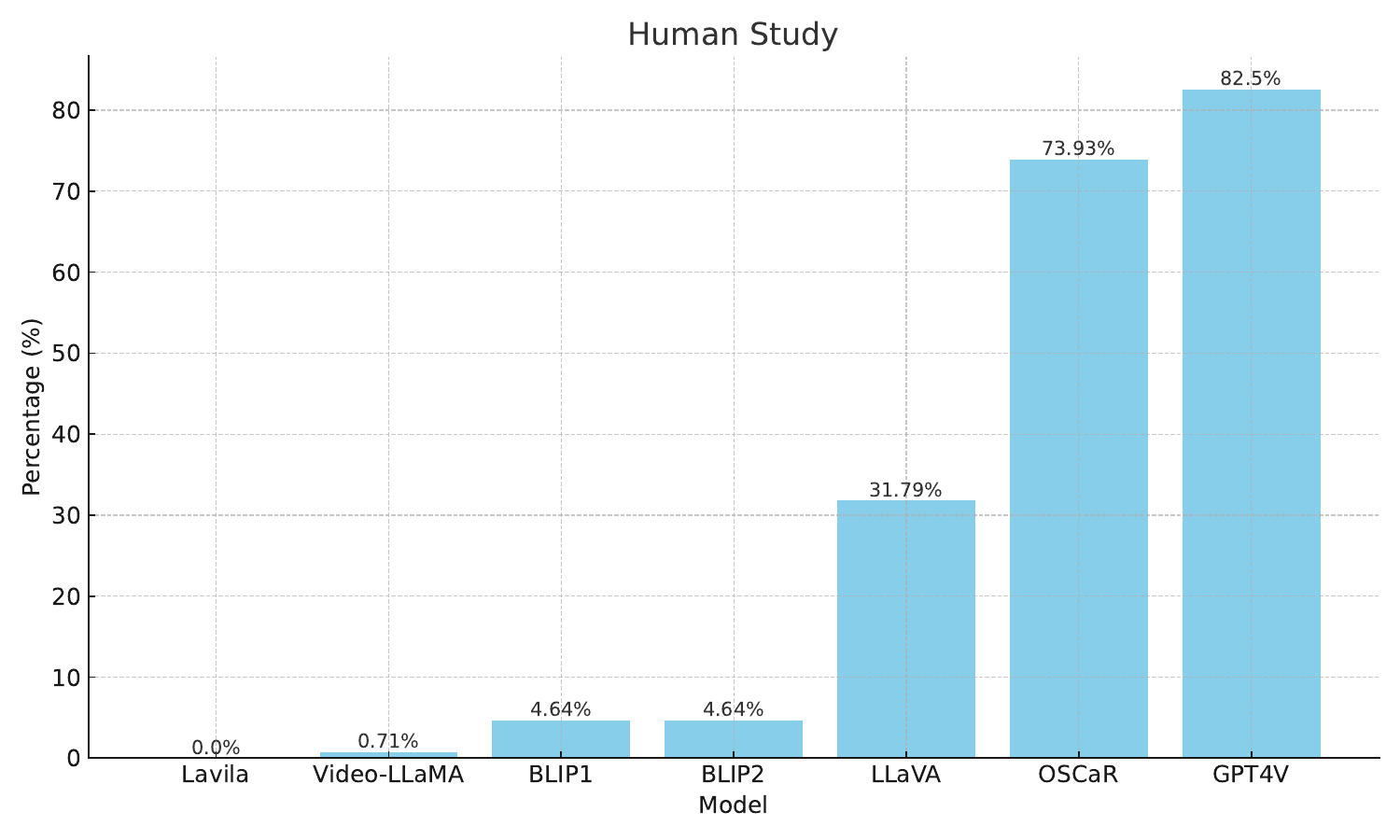}
\caption{\textbf{Human study results.} The figure shows the percentage that each model was selected by participants as producing favorable descriptions in a human rating study.}
\label{fig:human-study}
\end{figure}

\noindent \textbf{Human Study:} In our study to assess caption quality from various models, seven evaluators reviewed five videos with four captions each (three for frames, one for state changes), provided by seven models. Each caption had seven different options generated by seven different models. Evaluators could select up to two options per caption that they think are the best. Figure~\ref{fig:human-study} shows the results of this experiment. We calculated the percentage of times each model was selected and found that OSCaR achieved 73.93\%, which was only 8.57\% lower than GPT-4V. 
OSCaR significantly outperformed LLaVA by more than two times. These results demonstrate that OSCaR is a promising model for generating high-quality captions.
\subsection{Open-world Objects Evaluation}
\label{sec:open-world-eval}

Evaluating the performance of machine learning models solely based on objects seen during training isn't enough. To more thoroughly test their effectiveness, we also evaluated them on objects not included in the training set, representing the open world. In this part of our study, we compare the quality of text produced by our model and GPT-4V for these open-world objects, using BLEU and ROUGE scores as our metrics.

\noindent \textbf{In-domain Objects Evaluation:} The evaluation results on objects in the cooking domain that were not included in the instruction fine-tuning data are presented in Table~\ref{tab:openworld_performance}. When compared with the results in Table~\ref{tab:model_performance}, the overall performance is better when testing with in-domain open-world objects. One of the reasons for this is that the evaluation set in Table~\ref{tab:model_performance} was corrected by humans, while the data used in Table~\ref{tab:openworld_performance} was generated from GPT-4V. Nevertheless, the outcomes of this experiment indicate the generalizability of models when dealing with new objects. \\
\noindent \textbf{Objects Beyond Cooking Domain:} Table~\ref{tab:openworld_performance} presents the open-world evaluation for various domains. The dataset employed in this experiment is discussed in section~\ref{sec:open-world}, which comprises 356 videos from 51 distinct domains. Compared to the experiment in table~\ref{tab:model_performance}, the outcomes of this experiment are generally lower. Specifically, for LLaVA 7B with mixed data, this experiment shows a decline of 0.1, 0.95, 0.05, and 0.85 on BLEU, ROUGE-1, ROUGE-2, and ROUGE-L, respectively. This decline indicates two things: 1) the open domain is challenging and may require domain-specific data for fine-tuning to achieve better performance, and 2) even in the absence of new domain data, the decrease in performance is not too significant, and showing the generalizability of our model.\\ 

\subsection{Ablation Study}

\begin{table*}[t] 
\centering
\caption{\textbf{Performance comparison based on BLEU and ROUGE scores in different domains.} The table compares various models with open-world benchmarks.}
\label{tab:baseline_openworld}
\scalebox{0.7}{
\begin{tabularx}{\textwidth}{@{}lXcccc@{}}
\toprule
\textbf{Domain} & \textbf{Method} & \textbf{BLEU} & \textbf{ROUGE-1} & \textbf{ROUGE-2} & \textbf{ROUGE-L} \\
\midrule
\multirow{4}{*}{Cooking Domain} 
& LLaVA~\cite{Liu2023VisualIT} & 2.56 & 23.77 & 5.64 & 21.08 \\
& BLIP1~\cite{Li2022BLIPBL} & $4.33 \times 10^{-5}$ & 0.75 & 0.026 & 0.73 \\
& BLIP2~\cite{Li2023BLIP2BL} & 0.043 & 8.2 & 1 & 7.4 \\
& LaViLa~\cite{Zhao2022LearningVR} & $4.34 \times 10^{-5}$ & 3.09 & 0.27 & 3.07 \\
\midrule
\multirow{4}{*}{Other Domains} 
& LLaVA~\cite{Liu2023VisualIT} & 2.88 & 23.96 & 5.85 & 21.26 \\
& BLIP1~\cite{Li2022BLIPBL} & $7.39 \times 10^{-5}$ & 1.15 & 0.077 & 1.13 \\
& BLIP2~\cite{Li2023BLIP2BL} & 0.028 & 9.04 & 1.05 & 8.2 \\
& LaViLa~\cite{Zhao2022LearningVR} & $6.95 \times 10^{-5}$ & 3.1 & 0.29 & 3.07 \\
\bottomrule
\end{tabularx}
}
\end{table*}

Our research also examined the accuracy of video frame annotations in the EPIC-KITCHENS and Ego4D datasets. We used Amazon Mechanical Turk annotators to evaluate 500 video data points for the precision and completeness of descriptions, categorizing them into four classes. In addition, we analyzed 100 samples from each setting of zero-shot and two-shot to determine the best strategy for scaling up data annotation. Our findings indicate that zero-shot is the more effective approach for annotating our task's data.

Our findings, detailed in Table \ref{tab:2shots_vs_zeroshot}, compare zero-shot and two-shot performance in aligning descriptions with human standards of accuracy and relevance, as derived from video frame analyses. This table illustrates how well the GPT-4V model's natural language descriptions, evaluated by Amazon Mechanical Turk annotators in zero and two-shot scenarios, match human judgment. The percentages indicate the extent to which these descriptions accurately and relevantly depict the video content, based on a frame-by-frame review. Each description was judged for its thoroughness and relevance in detailing the object and its activities. Annotators followed established guidelines to determine the quality of data in their assessments.

The results reveal a notable disparity in description quality between the zero-shot and two-shot methods. The zero-shot approach yielded a higher proportion of Fully Detailed and Comprehensive descriptions, while the two-shots method indicated a greater occurrence of descriptions with errors or misleading content. This variation highlights the differences in data quality and annotator perceptions under varying evaluation conditions, underscoring the importance of method selection in annotation studies. 
\begin{table}[h]
\centering
\caption{The table lists the distribution of Amazon Mechanical Turk annotators' choices of descriptions of objects and object state changes in 0 and two-shot tests by the GPT-4V model in \%. }
\label{tab:2shots_vs_zeroshot}
\scalebox{0.8}{
\begin{tabularx}{1.1\columnwidth}{lXX}
\toprule
Satisfaction Class  & Zero-shot & Two-shots \\
\midrule
Fully Detailed & 56.25 & 33.25 \\
Minor Mistakes & 16.75 & 28.25 \\
Lacks Important Details & 13.25 & 23.00 \\
Hallucinating & 13.75 & 15.50 \\
\bottomrule
\end{tabularx}
}
\end{table}

In Table~\ref{tab:baseline_openworld}, we present the results of our experiment where we evaluate various models in open-world benchmarks, including the cooking domain and other domains. We have observed that the performance of other baselines has generally decreased in open-world benchmarks. These results demonstrate the importance of building models that can be generalized in the world. However, capturing the state of objects while dealing with diverse objects and domains is still a major challenge. 

\section{Conclusion}
This paper presents a new task for comprehending the state of objects and their changes using natural language. We also propose a data generation pipeline that utilizes the capabilities of GPT-4V to tackle this task. Furthermore, we introduce OSCaR, a dataset that includes training data and a benchmark with various protocols. Our comprehensive experiments not only demonstrate the superiority of our methods in comparison to previous state-of-the-art open-source solutions but also examine the limitations of GPT-4V in addressing this challenge.

\section{Limitations}
This study explores a new research problem that focuses on understanding the states of objects. Although it has provided valuable insights, some limitations and areas still require further investigation, as outlined below.

\noindent\textbf{Lack of audio integration:} A limitation of this work is the lack of audio data, which could be useful in scenarios where sound is essential for indicating changes or properties of objects.

\noindent \textbf{Challenges in long-term state transition tracking:} Tracking changes in object state over extended periods is challenging because many current models, especially foundation models and models based on LLMs, do not yet have the ability to capture long-term information. This limitation highlights the difficulty in understanding complex, long-term transitions in object states, which is critical to comprehending object dynamics in various environments.

\noindent \textbf{Reliance on GPT-4V's imperfect outputs:} Although GPT-4V has shown strength in generating data for this research problem, its outputs are imperfect. This limitation highlights the need for strategies to efficiently learn from and improve upon the imperfect data provided by GPT-4V.
\section{Ethics Statement}
We acknowledge that bias could be present in the process of collecting data for our paper. 
To minimize this issue, we have taken several measures. Firstly, we have collected videos from two highly diverse data sources: EPIC-KITCHENS and Ego4D. 
Secondly, when labeling the data using GPT-4V, we are aware that bias could occur from the behavior of GPT-4V. 
To address this, we regularly take test samples during the data generation process. If we detect any significant issues, we are prepared to stop the process and conduct an inspection. On the human side, we use the Amazon Mechanical Turk platform to hire people to label data for both the GPT4 zero-shot and few-shot quality assessment steps and the user study. Our data collection was classified as an approved exempt protocol by the IRB.

\section*{Acknowledgement}

This work was supported by NSF 2202124, NNSA NA0004078, NIH R01EY034562, and DARPA HR00112220003. The content of the information does not necessarily reflect the position of the Government, and no official endorsement should be inferred.

\bibliography{custom}

\appendix
\section{Appendix}
\subsection{Prompts}
\subsubsection{Data Generation}
Using a single request, we generated object state captions and state change captions for a single video with this prompt.
\newmdenv[
    leftmargin=10pt,
    rightmargin=10pt,
    backgroundcolor=gray!20,
    linecolor=black,
    linewidth=2pt,
    topline=false,
    bottomline=false,
    skipabove=\baselineskip,
    skipbelow=\baselineskip
]{quotebox}

\begin{quotebox}
These are three frames in a video a person <action> <object name>. Do these two tasks focusing on the <object name>.
1. Describe the state of the <object name> in these three frames in a detailed way separately. No need to tell me which frame you are describing, such as "In the first frame" or something like that. The captions should be very long with at least three to five sentences, very detail follow the task.
2. Describe the state change of the <object name> in these three frames in progress. The caption should be very long with at least three to five sentences, very detail followed the task.
The answer should be in the format of 4 paragraphs:

Frame\_1:

Frame\_2:

Frame\_3:

State\_change\_caption:
\end{quotebox}

\noindent The conversational data have been generated from this prompt:
\begin{quotebox}
<Caption here>

From this caption, give me a conversation with 10 pairs of question-answer focusing on the object’s state. The answer needs to be found in the caption, and the questions can also serve as a suggestion to generate this caption and not mention the caption in the question and answer. At least 3 questions and answers require reasoning ability. Output should follow the format
        Question\_1: Generate a question here
        
        Answer\_1: Generate the answer here (answer should be long, at least 3 sentences with an explanation)
        
        Question\_2: Generate a question here
        
        Answer\_2: Generate the answer here (answer should be long, at least 3 sentences with an explanation)
        
        Question\_3: Generate a question here
        
        Answer\_3: Generate the answer here (answer should be long, at least 3 sentences with an explanation)
        
        Question\_4: Generate a question here
        
        Answer\_4: Generate the answer here (answer should be long, at least 3 sentences with an explanation)
        
        Question\_5: Generate a question here
        
        Answer\_5: Generate the answer here (answer should be long, at least 3 sentences with an explanation)
        
        Question\_6: Generate a question here
        
        Answer\_6: Generate the answer here (answer should be long, at least 3 sentences with an explanation)
        
        Question\_7: Generate a question here
        
        Answer\_7: Generate the answer here (answer should be long, at least 3 sentences with an explanation)
        
        Question\_8: Generate a question here
        
        Answer\_8: Generate the answer here (answer should be long, at least 3 sentences with an explanation)
        
        Question\_9: Generate a question here
        
        Answer\_9: Generate the answer here (answer should be long, at least 3 sentences with an explanation)
        
        Question\_10: Generate a question here
        
        Answer\_10: Generate the answer here (answer should be long, at least 3 sentences with an explanation)

\end{quotebox}

\noindent We used this prompt to generate multiple-choice question-answer pairs:
\begin{quotebox}
    From this caption, give me 10 multiple-choice questions focusing on the object's state. The answer needs to be found in the caption, and the questions can also serve as a suggestion to generate this caption and do not mention the caption in question.
    
<Caption here>
\end{quotebox}

\subsubsection{Evaluation}

In addition to standard text generation metrics, we evaluated output using GPT-4V. To do this, we used the below prompt for evaluation. For example, if we need to evaluate GPT-4V, OSCaR, and LLaVA, we use the below prompt.

\begin{quotebox}
    This is an image of a human interacting with <object name>. I want you to do this task: Rate the answers, including GPT4, on a scale from 0 to 100. 100 is perfect. N/A will also be 0. Also, provide an explanation about your rate. The answer need to follow the format:
    
Rate\_GPT4\_accuracy: rate for accuracy of GPT4 answer, only give a number from 0 to 100.

Rate\_GPT4\_helpfulness: rate for helpfulness GPT4 answer, only give a number from 0 to 100.

Rate\_GPT4\_detaillevel: rate for detail level of GPT4 answer, only give a number from 0 to 100.

Rate\_GPT4\_reasoning: rate for reasoning ability of GPT4 answer, only give a number from 0 to 100.

Rate\_GPT4\_overall: rate for overall of GPT4 answer, only give a number from 0 to 100.

Rate\_OSCaR\_accuracy: rate for accuracy of OSCaR answer, only give a number from 0 to 100.

Rate\_OSCaR\_helpfulness: rate for the helpfulness of OSCaR answer, only give a number from 0 to 100.

Rate\_OSCaR\_detaillevel: rate for detail level of OSCaR answer, only give a number from 0 to 100.

Rate\_OSCaR\_reasoning: rate for reasoning ability of OSCaR answer, only give a number from 0 to 100.

Rate\_OSCaR\_overall: rate for overall of OSCaR answer, only give a number from 0 to 100.

Rate\_LLaVA\_accuracy: rate for accuracy of LLaVA answer, only give a number from 0 to 100.

Rate\_LLaVA\_helpfulness: rate for the helpfulness of LLaVA answer, only give a number from 0 to 100.

Rate\_LLaVA\_detaillevel: rate for detail level of LLaVA answer, only give a number from 0 to 100.

Rate\_LLaVA\_reasoning: rate for reasoning ability of LLaVA answer, only give a number from 0 to 100.

Rate\_LLaVA\_overall: rate for overall of LLaVA answer, only give a number from 0 to 100.

Here is the question and list of answers:

Question: <question>

GPT4: <GPT4 answer>

OSCaR: <OSCaR answer>

LLaVA: <LLaVA answer>
\end{quotebox}

\end{document}